\documentclass{article}

\usepackage{times}
\usepackage{graphicx} 
\usepackage{subfigure} 

\usepackage{natbib}

\usepackage{algorithm}
\usepackage{algorithmic}

\usepackage{hyperref,array,multirow}
\usepackage{pbox}
\usepackage{natbib,url}
\usepackage{color,graphicx,amsmath,amsthm,amssymb,amsbsy,amsfonts,booktabs,multirow}
\usepackage{enumitem}

\newcolumntype{C}[1]{>{\centering\let\newline\\\arraybackslash\hspace{0pt}}m{#1}}

\newtheorem{thm}{Theorem}
\newtheorem{defn}[thm]{Definition}

\usepackage[accepted]{icml2015}

\icmltitlerunning{Empirical Study of SVI for the Beta Bernoulli Process}

\begin{document} 

\twocolumn[
\icmltitle{An Empirical Study of Stochastic Variational \\ Algorithms 
           for the Beta Bernoulli Process}

\icmlauthor{Amar Shah $^\star$}{as793@cam.ac.uk}
\icmlauthor{David A. Knowles $^\dagger$}{davidknowles@cs.stanford.edu}
\icmlauthor{Zoubin Ghahramani $^\star$}{zoubin@eng.cam.ac.uk}
\icmladdress{$^\star$ University of Cambridge, Department of Engineering, Cambridge, UK}
\vspace{-3mm}
\icmladdress{$^\dagger$ Stanford University, Department of Computer Science, Stanford, CA, USA}

\icmlkeywords{boring formatting information, machine learning, ICML}

\vskip 0.175in
]

\begin{abstract} 
Stochastic variational inference (SVI) is emerging as the most promising candidate for scaling inference in Bayesian probabilistic models to large datasets. However, the performance of these methods has been assessed primarily in the context of Bayesian topic models, particularly latent Dirichlet allocation (LDA). Deriving several new algorithms, and using synthetic, image and genomic datasets, we investigate whether the understanding gleaned from LDA applies in the setting of sparse latent factor models, specifically beta process factor analysis (BPFA). We demonstrate that the big picture is consistent: using Gibbs sampling within SVI to maintain certain posterior dependencies is extremely effective. However, we find that different posterior dependencies are important in BPFA relative to LDA. Particularly, approximations able to model
intra-local variable dependence perform best. 
\end{abstract} 

\section{Introduction}

The last two decades have seen an explosion in the development of flexible statistical methods able to model diverse data sources. Bayesian nonparametric priors in particular provide a powerful framework to enable models to adapt their complexity to the data at hand~\citep{orbanz2010bayesian}. In the regression setting this might mean learning the smoothness of the output function~\citep{rasmussen2006gaussian}, for clustering adapting the number of components~\citep{maceachern1998estimating}, and in the case of our interest, latent factor models, finding an appropriate number of latent features~\citep{knowles2011nonparametric}. While such models are appealing for a range of applied data analysis applications, their scalability is often limited. The posterior distribution over parameters and latent variables is typically analytically intractable and highly multimodal, making MCMC, particularly Gibbs sampling, the norm. Along with concerns over performance and convergence, MCMC methods are often impractical for the applied practitioner: how should the multiple samples be summarized? Variational methods work on the basis that simply finding a good posterior mode, and giving some measure of the associated uncertainty, is typically sufficient. In addition, the predictive performance of variational methods is often comparable to more computationally expensive sampling based approaches \citep{ghahramani1999}. 

Recently \emph{stochastic} variational inference has begun to emerge as the most promising avenue for scaling inference in large latent variable models \citep{svi}. Marrying variational inference with stochastic gradient descent allows principled updates using only minibatches of observations, greatly improving data scalability. While some MCMC methods have been proposed to work with mini-batches \citep{welling2011bayesian,ahn2012bayesian} they lack theoretical guarantees and apply only to continuous, unbounded latent variables. While SVI has been influential for Bayesian topic modeling, particularly latent Dirichlet allocation \citep[LDA,][]{mimno2012,ssvi,wang2012truncation}, the same cannot be said for sparse factor analysis models for continuous data. While the former has been driven by the ready availability of huge text corpora, the scale of continuous data being generated by new genomic technologies is still growing. For example, CyTOF, single cell time of flight mass spectrometry is able to measure the abundance of dozens of proteins in hundreds of thousands of cells in a single run \citep{bendall2011single}. Such complex, large scale, high dimensional datasets require sophisticated statistical models, but are typically analyzed using simple heuristic clustering methods or PCA, which do not capture important structure, such as sparsity. As a result, scaling more advanced factor analysis type models is of great interest. 

When desigining a variational approximation to a posterior distribution, one must trade off the accuracy of the approximation with the complexity of optimizing the evidence lower bound. Mean field approximations are simple to work with, but \citet{ssvi} demonstrated that in the context of LDA, maintaining posterior dependence between ``global'' variables (topic vectors) and ``local'' variables (document vectors) is crucial to finding good solutions. Does this finding hold for sparse factor analysis models? Our results suggest that contrary to the LDA case, maintaining dependencies \emph{amongst local variables} is actually the most important ingredient for obtaining good performance with beta Bernoulli process SVI. 


\section{Beta Process for Factor Analysis}

The \textit{beta process} \citep{hjort,thib07} is an independent increments process  defined as follows:

\begin{defn}
Let $\Omega$ be a measurable space and $\mathcal{B}$ its $\sigma$-algebra. 
Let $H_0$ be a continuous probability measure on $(\Omega, \mathcal{B})$
and $\alpha$ a positive scalar. Then for all disjoint, infinitesimal partitions, 
$\{B_1,...,B_K\}$, of $\Omega$ the beta process is generated as follows,
\begin{equation}
H(B_k) \stackrel{\mathrm{iid}}{\sim} \mathrm{Beta}( \alpha H_0(B_k), \alpha (1 - H_0(B_k) ) )
\end{equation}
with $K \rightarrow \infty$ and $H_0(B_k)\rightarrow 0$ for $k=1,...,K$. We 
denote the process $H \sim \mathrm{BP}(\alpha H_0)$.
\end{defn} 

Hjort considers a generalization of this definition including functions, $\alpha(B_k)$,
which we set as constants for the sake of simplicity. Analogous to the Dirichlet process, 
the beta process may be written in set function form as
\begin{equation}
H( \omega) = \sum_{k=1}^\infty \pi_k \delta_{\omega_k} (\omega)
\label{bp}
\end{equation} 
with $H(\omega_i) = \pi_i$. Note that the beta process is not a normalized random
measure. Hence the $\boldsymbol{\pi}$ of a 
beta process does not represent a probability mass function on $\Omega$, but 
instead can be used to parametrize the \textit{Bernoulli process}, a new measure on $\Omega$ defined
as follows:

\begin{defn}
Let $\boldsymbol{z_i}$ be an infinite row vector with $k^\mathrm{th}$ value, $z_{ik}$,
generated by $z_{ik}|\pi_k \stackrel{\mathrm{iid}}{\sim} \mathrm{Bernoulli}(\pi_k)$.
The measure defined by $X_i(\omega) = \sum_k z_{ik}\delta_{\omega_k}(\omega)$ is
then a draw from a Bernoulli process, which we denote $X_i \sim \mathrm{BeP}(H)$. 
\end{defn}

If we were to stack samples of the infinite-dimensional vector, $\boldsymbol{z_i}$, to form a matrix, 
$\boldsymbol{Z} = [\boldsymbol{z_1}^\top,...,\boldsymbol{z_N}^\top]^\top$, we may view the beta-Bernoulli process as a prior over infinite binary matrices \citep{grif5}, where each column in the matrix $\boldsymbol{Z}$ corresponds to a location, $\delta_\omega$.

Sampling $H$ directly, as defined in (\ref{bp}), is difficult to do exactly and 
efficiently. But, just as \citet{aldous} derived the Chinese restaurant process, a marginalized
approach used for sampling from the Dirichlet process, there exists an efficient marginalized
scheme for sampling from the beta process, called the Indian buffet process  \citep[IBP,][]{grifnips,thib07}.  

The IBP sampling procedure introduces strong dependencies 
between the rows of $\boldsymbol{Z}$. Our goal is to derive a stochastic
variational inference scheme where we consider rows in batches. It will hence be crucial
to instantiate the global parameters rather than marginalize over them.

For this reason, we shall consider a finite approximation to the beta process which simply
set $K$ to a large, finite number. The finite representation is written as
\begin{align}
H(\omega) &= \sum_{k=1}^K \pi_k \delta_{\omega_k}(\omega) \notag \\
\pi_k \sim \mathrm{Beta}(a/K,b&(K-1)/K) , \hspace{6mm}
\omega_k \sim H_0
\label{trunc}
\end{align}
and the $K$-dimensional vector, $\boldsymbol{z_i}$, is drawn from a finite Bernoulli process parameterized by 
$H$. 

Consider modelling a data matrix $\boldsymbol{Y} \in \mathbb{R}^{N \times D}$ where rows represent data points. Factor analysis models this data as the product of two matrices
$\boldsymbol{L} \in \mathbb{R}^{N \times K}$ and $\boldsymbol{\Phi} \in \mathbb{R}^{K \times D}$, plus an error matrix, $\boldsymbol{E}$. 
\begin{equation}
\boldsymbol{Y} = \boldsymbol{L\Phi} + \boldsymbol{E}
\end{equation} 
Prior belief about the structure of the data may be used to induce the desired propeties of $\boldsymbol{L}$ and 
$\boldsymbol{\Phi}$, e.g. sparsity \citep{west,ihfrm,knowles}. To encourage sparsity, we model $\boldsymbol{L}$ as the Hadamard (element-wise) product between matrices $\boldsymbol{Z}$ and $\boldsymbol{W}$, $\boldsymbol{L} = \boldsymbol{Z} \circ \boldsymbol{W}$,
where $\boldsymbol{Z}$ is binary and $\boldsymbol{W}$ is a Gaussian weight matrix. This idea is described in Section 3 of \citep{grif5}. We model the matrices $\boldsymbol{\Phi}$ and $\boldsymbol{Z}$ as $N$ draws from a beta-Bernoulli process parameterized 
by a beta process, $H$. 

Using the truncated beta process of (\ref{trunc}), we have the following 
generative process for observation $i = 1,...,N$ and features $k=1,...,K$,  
\begin{align}
\label{generative}
\boldsymbol{y_i} &= (\boldsymbol{z_i} \circ \boldsymbol{w_i} )\boldsymbol{\Phi} + \boldsymbol{\epsilon_i} \\
\boldsymbol{w_i} &\sim \mathcal{N}(0, \gamma^{-1}_w I ) \notag \\
z_{ik}|\pi_k &\sim \mathrm{Bernoulli}(\pi_k) \notag \\
\boldsymbol{\epsilon_i} &\sim \mathcal{N}(0,\gamma_\mathrm{obs}^{-1} I )  
          \hspace{23mm} \smash{\raisebox{\dimexpr1.0\normalbaselineskip+1.1\jot}{$%
            \left.\begin{array}{@{}c@{}}\\[\jot]\\[\jot]\\[\jot]\end{array}\right\}\text{Local variables}
          $}}\notag \\
\pi_k &\sim \mathrm{Beta}(a/K, b(K-1)/K) \notag \\
\boldsymbol{\phi_k} &\sim \mathcal{N}(0, D^{-1} \boldsymbol{I}) \notag 
    \hspace{22mm} \smash{\raisebox{\dimexpr0.6\normalbaselineskip+0.8\jot}{$%
            \left.\begin{array}{@{}c@{}}\\[\jot]\\[\jot]\end{array}\right\}\text{Global variables}
          $}} 
\end{align}
where all values are drawn independently. This is the generative model used for beta process factor
analysis \citep{bpfa}. We place independent Gamma$(c',d')$ and Gamma$(e',f')$ priors on $\gamma_\mathrm{obs}$ and $\gamma_w$ respectively. The separation of local and global variables will be crucial for the stochastic variational inference algorithm which we derive in the next section. For the sake of brevity, we denote the set of global variables
$\boldsymbol{\beta} \equiv \{\boldsymbol{\pi},\boldsymbol{\Phi},\gamma_w, \gamma_\mathrm{obs} \}$ and sets of local variables $\boldsymbol{\psi}_i \equiv \{\boldsymbol{w}_i,\boldsymbol{z}_i \}$ for $i=1,...,N$.

\section{Variational Inference Schemes}

The true posterior distribution $p(\boldsymbol{\beta},\boldsymbol{\psi}_{1:N}|\boldsymbol{x}_{1:N})$
involves complicated dependencies between latent variables, which makes inference complicated. The goal of variational inference is to approximate the true posterior with a family of distributions $q(\boldsymbol{\beta},\boldsymbol{\psi}_{1:N})$. We choose the best member of the chosen family of distributions by minimizing the KL-divergence between this variational distribution and the true posterior. Equivalently, we maximize the \textit{evidence lower bound} (ELBO),
\begin{equation}
\mathcal{L}(q) = \mathbb{E}_q[ \log p(\boldsymbol{\beta},\boldsymbol{\psi}_{1:N},\boldsymbol{x}_{1:N})- \log q(\boldsymbol{\beta},\boldsymbol{\psi}_{1:N})].
\end{equation} 
In this work, we compare the performance of a range of variational approximations. Each of the approximations we consider factorizes as follows 
\begin{equation*}
q(\boldsymbol{\beta},\boldsymbol{\psi}_{1:N}) = \Big(
 q(\gamma_\mathrm{obs}) q(\gamma_w) \prod_k q(\pi_k) q(\boldsymbol{\phi}_k) \Big)  q(\boldsymbol{\psi}_{1:N} | \boldsymbol{\beta}),
\end{equation*}
where $q(\gamma_\mathrm{obs}) = \mathrm{Gamma}(c,d)$, 
$q(\gamma_w) = \mathrm{Gamma}(e,f)$,
$q(\boldsymbol{\phi}_k) = \mathcal{N}(\tau_k^{-1} \boldsymbol{\mu}_k, \tau_k^{-1} \boldsymbol{I})$ and 
$q(\pi_k) = \mathrm{Beta}(a_k,b_k)$. The global variational distributions are all of the same exponential family forms as their posterior conditional distributions. 
The full set of global variational parameters is $\boldsymbol{\lambda} = \{ a_k, b_k, c,d,e,f, \tau_k, \boldsymbol{\mu}_k \}$. Due to conjugacy of our model, it is easy to show that the updates for the global variational parameters during the variational M-step are as follows
\begin{align}
\label{global updates}
a_k &= a/K + \sum_i \mathbb{E}_q[z_{ik}]   \\
b_k &= b(K-1)/K + \sum_i \Big(1-\mathbb{E}_q[z_{ik}] \Big) \notag \\
c &= c' + \sum_i D/2 \notag \\
d &= d' +  \sum_i \frac{1}{2} \mathbb{E}_q \Big[ \big\| \boldsymbol{y}_i -  (\boldsymbol{z_i} \circ \boldsymbol{w_i} )\boldsymbol{\Phi} \|^2 \Big] \notag \\
e &= e' + \sum_i K/2 \notag \\
f &= f' +  \sum_i \frac{1}{2} \mathbb{E}_q \big[ \boldsymbol{w_i} \boldsymbol{w_i}^\top \big] \notag \\
\tau_k &= D + \sum_i \mathbb{E}_q \big[\gamma_\mathrm{obs} z_{ik} {w_{ik}}^2 \big] \notag \\
\boldsymbol{\mu}_k &= \sum_i \mathbb{E}_q \big[ z_{ik} w_{ik} \boldsymbol{y}_i^{-k} \big] \notag
\end{align}
where $\mathbb{E}_q$ is an expectation over all latent variables with respect to $q$ (except when global parameters are being sampled), and $\boldsymbol{y}_i^{-k} =  \boldsymbol{y}_i - \mathbb{E}_q \big[ \sum_{j \neq k} z_{ij} w_{ij} \boldsymbol{\phi}_j \big]$. We work with the \textit{natural parameters} of the global variational distributions. Natural gradients give the direction of steepest ascent in Riemannian space, leading to faster convergence for e.g. maximum likelihood estimation \citep{amari}. 

Our aim is to update the global variational parameters stochastically, by considering subsets of the full dataset and making sequential updates. Let $\boldsymbol{\eta}$ denote the vector of global natural parameters of $p(\boldsymbol{\beta})$, and by conditional conjugacy, the vector of global natural parameters of $q(\boldsymbol{\beta})$ is $\boldsymbol{\eta} + \sum_i \boldsymbol{\eta}_i(\boldsymbol{y}_i,\boldsymbol{\psi}_i)$. In fact, $\boldsymbol{\eta} = [a/K,b(K-1)/K,c',d',e',f',D,\boldsymbol{0}]$, and
$\boldsymbol{\eta}_i$ is a vector consisting of each of the $i^{th}$ elements of the sums in Equation \ref{global updates}.  We have followed the notation of \citet{ssvi}. 
The general framework of stochastic variational inference we shall follow is summarized in Algorithm \ref{algo}.

The difficult step is in computing $\hat{\boldsymbol{\eta}}_i$, and it is entirely dependent on the form of the local variable approximation, of which we consider 2 types: `Unstructured' methods where 
$q(\boldsymbol{\psi}_{1:N}|\boldsymbol{\beta}) = q(\boldsymbol{\psi}_{1:N})$ and `structured' methods where this equivalence does not hold. Our notion of `structure' describes the dependence between local and global variables, as discussed by \citet{ssvi}.  

\subsection{Unstructured Variational Methods}

The simplest, and most commonly used, approximation we can make is the mean field approximation, 
\begin{align}
q_\mathrm{MF}(\boldsymbol{\psi}_i) =  \prod_{k=1}^K q(z_{ik}) q(w_{ik}) 
\end{align}
where $q(z_{ik}) = \mathrm{Bernoulli}(\theta_{ik})$ and $q(w_{ik})= \mathcal{N}(\kappa_{ik}^{-1}\nu_{ik},\kappa_{ik}^{-1})$. Given the current set of global parameters, $\boldsymbol{\lambda}^{(t)}$, the local ELBO, 
$\mathcal{L}_\mathrm{local}= \mathbb{E}_{q(\boldsymbol{\beta})q_\mathrm{MF}(\boldsymbol{\psi}_{1:N})}[\log p(\boldsymbol{y}_{1:N},\boldsymbol{\psi}_{1:N}|\boldsymbol{\beta}) - \log q(\boldsymbol{\psi}_{1:N})]$ is optimized as a function of local variational parameters $\{\theta_{ik},\nu_{ik},\kappa_{ik} \}$. Up to irrelevant constants, 
\begin{align}
\mathcal{L}_\mathrm{local}^\mathrm{MF-SVI} &= \frac{c}{2d} \sum_{i,k} \bigg[ 2\theta_{ik} \frac{\nu_{ik}}{\kappa_{ik}} \frac{\boldsymbol{\mu}_k}{\tau_k} \boldsymbol{y}_i^\top \notag \\
&\hspace{-8mm} - \theta_{ik}\bigg(\frac{{\nu_{ik}}^2}{{\kappa_{ik}}^2} + \frac{1}{\kappa_{ik}} \bigg)
\bigg(\frac{\boldsymbol{\mu}_k \boldsymbol{\mu}_k^\top}{{\tau_k}^2} + \frac{1}{\tau_k}\bigg) \notag \\
&\hspace{-8mm} - \sum_{j \neq k} \theta_{ij}\theta_{ik} \frac{\nu_{ij}}{\kappa_{ij}} \frac{\nu_{ik}}{\kappa_{ik}}
\frac{\boldsymbol{\mu}_j \boldsymbol{\mu}_k^\top}{\tau_j \tau_k} \bigg] - \frac{e}{2f} \sum_{i,k}
\bigg(\frac{{\nu_{ik}}^2}{{\kappa_{ik}}^2} + \frac{1}{\kappa_{ik}} \bigg)  \notag \\
&\hspace{-8mm} + \sum_{i,k} \theta_{ik} (\psi(a_k)-\psi(b_k)) - \frac{1}{2} \sum_{i,k} \log (\kappa_{ik}) \notag \\
&\hspace{-8mm} - \sum_{i,k} \big[ \theta_{ik} \log \theta_{ik} + (1-\theta_{ik}) \log (1-\theta_{ik})  \big]
\end{align}
where $\psi$ is the digamma function. The mean field approximation breaks dependencies between all local and global variables, and will provide a baseline to compare against. 
It is possible to compute
$\mathbb{E}_{q_\mathrm{MF} (\boldsymbol{\psi}_i)}[\boldsymbol{\eta}_i]$ analytically given the optimized local variational parameters. We denote the SVI algorithm which uses a mean field local variable approximation as MF-SVI. It is identical to the original SVI algorithm introduced by \citet{svi}.

\citet{mimno2012} suggested an online SVI method which maintains structure between local variables specifically for the LDA. We generalize their idea by suggesting the following variational distribution over local parameters
\begin{align}
q_\mathrm{Mimno}(\boldsymbol{\psi}_i) = \exp( \mathbb{E}_{q(\boldsymbol{\beta})}[\log p(\boldsymbol{\psi}_i|\boldsymbol{y}_{1:N},\boldsymbol{\beta})] ) 
\end{align}
where $p(\boldsymbol{\psi}_i|\boldsymbol{y}_{1:N},\boldsymbol{\beta})$ is the true posterior conditional of $\boldsymbol{\psi}_i$. Whilst we are unable to compute $\mathbb{E}_{q_\mathrm{Mimno} (\boldsymbol{\psi}_i) }[\boldsymbol{\eta}_i]$ analytically, we are able to estimate it using MCMC. The SVI algorithm using $q_\mathrm{Mimno}$ as the local variational distribution shall be called Mimno-SVI. 

\subsection{Structured Variational Methods} 

Instead of taking an expectation over $q(\boldsymbol{\beta})$ to compute $\hat{\boldsymbol{\eta}}_i$ as in the previous section, we use the current set of global parameters $\boldsymbol{\lambda}^{(t)}$ to draw a sample $\boldsymbol{\beta}^{(t)}$, and compute an estimate of 
$\mathbb{E}_{q(\boldsymbol{\psi}_i | \boldsymbol{\beta}^{(t)})}[\boldsymbol{\eta}_i]$. Under this framework, Algorithm \ref{algo} becomes the SSVI-A algorithm of \citet{ssvi}.

Once again, the simplest approximation that can be made is the conditional mean-field approximation, where $z_{ik}, w_{ik}$ are independent given $\boldsymbol{\beta}$, with $q(z_{ik}) = \mathrm{Bernoulli}(\theta_{ik})$ and $q(w_{ik})= \mathcal{N}(\kappa_{ik}^{-1}\nu_{ik},\kappa_{ik}^{-1})$. This time, we optimize the local ELBO, $\mathbb{E}_{q_\mathrm{MF}(\boldsymbol{\psi}_{1:N}|\boldsymbol{\beta}^{(t)})}[\log p(\boldsymbol{y}_{1:N},\boldsymbol{\psi}_{1:N}|\boldsymbol{\beta}^{(t)}) - \log q(\boldsymbol{\psi}_{1:N})]$ as a function of local variational parameters $\{\theta_{ik},\nu_{ik},\kappa_{ik} \}$, and compute $\mathbb{E}_{q_\mathrm{MF}(\boldsymbol{\psi}_{1:N}|\boldsymbol{\beta}^{(t)})}[\boldsymbol{\eta}_i]$ analytically given the optimized parameters. We shall call this SVI method MF-SSVI.

The Bernoulli-Gaussian products present in the generative process in Equation \ref{generative} can be thought of as a spike-and-slab model. \citet{titsias2011} developed a variational method which maintains dependence between $z_{ik}$ and $w_{ik}$ for eack $k$, such that
\begin{align}
q_\mathrm{Titsias}&(\boldsymbol{\psi}_i|\boldsymbol{\beta}^{(t)}) = \prod_k 
\mathrm{Bernoulli}( z_{ik} ; \theta_{ik} ) \\
&\times 
\mathcal{N}\big( w_{ik} ; z_{ik} \kappa^{-1}_{ik} \nu_{ik}, z_{ik} \kappa^{-1}_{ik} + (1-z_{ik}) {\gamma_w^{(t)}}^{-1} \big). \notag
\end{align}
This approximation has the advantage that it maintains the spike-slab beaviour of the product $z_{ik}w_{ik}$, and matches the exact posterior when $z_{ik}=0$. However, the dependencies between local variables for which $k \neq k'$ are lost. Analogous to MF-SSVI, we optimize the local ELBO using $q_\mathrm{Titsias}$ as a function of $\{\theta_{ik},\nu_{ik},\kappa_{ik} \}$, and compute $\mathbb{E}_{q_\mathrm{Titsias}(\boldsymbol{\psi}_{1:N}|\boldsymbol{\beta}^{(t)})}[\boldsymbol{\eta}_i]$ analytically given the optimized parameters. We denote the SVI algorithm which uses the \citet{titsias2011} local approximation as Titsias-SSVI.

Finally we consider using the exact local conditional distribution given by $q(\boldsymbol{\psi}_i|\boldsymbol{\beta}^{(t)}) = p(\boldsymbol{\psi}_i|\boldsymbol{\beta}^{(t)},\boldsymbol{y}_i)$. We use MCMC samples to compute $\hat{\boldsymbol{\eta}}_i$ using a Gibbs sampling scheme. We therefore call this method Gibbs-SSVI.

MF-SVI \citep{svi}, MF-SSVI, Gibbs-SSVI \citep{ssvi} and Mimno-SVI \citep{mimno2012} have been considered in the context of LDA before, but the latter 3 have not been applied to factor analysis to the best of our knowledge. Titsias-SSVI is a new method as \citet{titsias2011} applied their variational approximation only to regression tasks. More details on the variational approximations over local variables is provided in the appendix. 

\begin{algorithm}[tb]
   \caption{General Stochastic Variational Inference}
   \label{algo}
\begin{algorithmic}
   \STATE Initialize $t=1$, $\boldsymbol{\lambda}^{(0)}$.
   \REPEAT
   \STATE Compute step size $\rho^{(t)} = (t + t_0)^{-\zeta}$.
   \STATE Select subset of full data set, $\mathcal{D}$.
   \STATE Compute $\hat{\boldsymbol{\eta}}_i$, an (unbiased) estimator of $\mathbb{E}_{q(\boldsymbol{\psi}_i | \boldsymbol{\beta})}[\boldsymbol{\eta}_i]$ for each $i  \in \mathcal{D}$   	   
\STATE Set $\boldsymbol{\lambda}^{(t)} = (1-\rho^{(t)}) \boldsymbol{\lambda}^{(t-1)} + \rho^{(t)} \big(\boldsymbol{\eta} + \frac{N}{| \mathcal{D}|} \sum_{i \in \mathcal{D}} \hat{\boldsymbol{\eta}}_i \big)$
   \UNTIL{convergence}
\end{algorithmic}
\end{algorithm}

\section{Related Work}

The idea of applying variational inference to the Indian buffet process was first proposed in \citet{doshi2008variational}, based on the stick breaking construction of the IBP \citep{teh7}. Promising results were shown for the simple but somewhat limited ``linear Gaussian'' model, which is the model presented here without the weight vector, $\boldsymbol{w}_i$.  \citet{bpfa} consider the simpler finite approximation to the beta process described above, and extended the model to include continuous weights $w_i$. An extension using power-EP, able to handle non-negativity constraints, was developed in \citep{ding2010nonparametric} but has not been widely adopted. Alternative approaches to scale inference in IBP based models have included parallelization \citep{doshi2009large} and submodular optimization \citep{ghahramani2013scaling}. The former only performed approximate sampling, and the later is greedy and limited to positive weights. Mean field based stochastic variational inference schemes have been used for large scale dictionary learning, with some success \citep{li2012,polatkan2014}. However, we shall show that preserving dependencies between local variables will greatly improves performance on image interpolation and denoising tasks. 

Meanwhile the topic modelling community has taken great strides developing stochastic variational inference methods for latent Dirichlet allocation \citep{blei2003latent}, encouraged by the availability of large corpora of text. The idea was initially proposed in \citet{hoffman2010online}, and refined in \citet{mimno2012} where the sparse updates of Gibbs sampling were leveraged to scale inference on just a single machine to 1.2 million books. The latter idea allows non-truncated online learning \citep{wang2012truncation} of Bayesian non-parametric models, though only the hierachical Dirichlet process \citep{teh2004sharing} was demonstrated. 

More recently, \citet{ssvi,liang2014} have shown that sampling from the global variational distribution improves predictive performance for the LDA and Bayesian non-negative matrix factorization respectively. In fact, the idea of optimizing an intractable variational inference algorithm by sampling from global variational distributions has been proposed in various contexts to deal with non-conjugacy \citep{ji2010,nott2012,gerrish,paisley2012,ranganath2014}. \citet{kingma2014,titsias2014,salimans2013} propose change of variable methods to deal with non-conjugacy or improve convergence speed. In this work we focus more on the quality of the variational approximation and attempt to exploit the conditional conjugacy. 

\begin{figure}[t]
    \includegraphics[trim = 25mm 185mm 117mm 19.5mm, clip,width=0.7\linewidth]{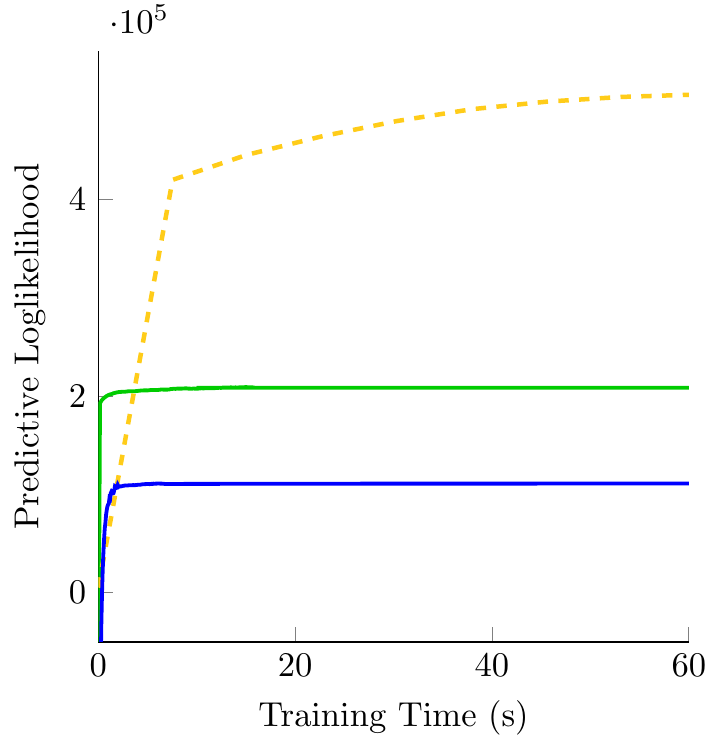}
\hspace{-4mm}
\raisebox{0.77\height}{\includegraphics[trim = 97mm 213mm 90mm 38mm, clip,width=0.32\linewidth]{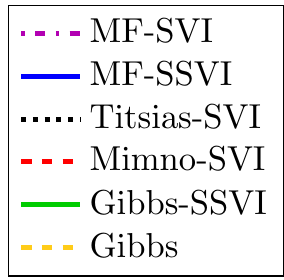}}    
\caption{Predicitve loglikelihood versus training time on synthetically generated data, comparing Gibbs-SSVI, MF-SSVI and Gibbs sampling. The same legend is used throughout this paper.}
\label{gibbsvsSVI}
\end{figure}

\section{Experiments}

In this section we discuss our findings from a range of experiments.
Results from experiments carried out on synthetically generated data are discussed first. We apply a range of stochastic variational inference algorithms to carry out image inpainting and denoising tasks next. Finally the same algorithms are applied to two large genomic datasets. We choose to compare our models using predictive loglikelihood of held out data, which we compute as

\begin{align}
p \Big(\boldsymbol{\hat{Y}}|\boldsymbol{Y} \Big) & \approx
\int p(\boldsymbol{\hat{Y}}|\boldsymbol{\beta},\boldsymbol{\psi}_{1:N_\mathrm{test}})
q(\boldsymbol{\beta},\boldsymbol{\psi}_{1:N_\mathrm{test}})
d(\boldsymbol{\beta},\boldsymbol{\psi}_{1:N_\mathrm{test}})
\notag \\
& \hspace{-10mm} \approx
\frac{1}{M} \sum_{m=1}^M \sum_{i=1}^{N_{\mathrm{test}}} 
\mathcal{N} \Big( \boldsymbol{\hat{y}}_n | \boldsymbol{z}_i^{(m)} \circ 
\boldsymbol{w}_i^{(m)} \boldsymbol{A}^{(m)} , \boldsymbol{I}/\gamma_{\mathrm{obs}}^{(m)} \Big)
\end{align}
where $\big(\boldsymbol{z}_i^{(m)}, \boldsymbol{w}_i^{(m)}, \boldsymbol{A}^{(m)}, \gamma_\mathrm{obs}^{(m)} \big)$ are independent samples from $q$, for whichever type of variational approximation is being used, and $\boldsymbol{\hat{y}}_i$ is the $i^{th}$ data point in the test set. 

In each of our experiments, we transform the data to have empirical mean 0 and variance 1. Hyperparameters are set as follows: $a = b = 10$, $c = 1$, $d=10$, $e=f=1$, and a learning rate schedule of $\rho_t = t^{-0.75}$ is employed. 

\begin{figure}[!t]
  \centering
    \includegraphics[trim = 25mm 185mm 117mm 20mm, clip,width=0.715\linewidth]{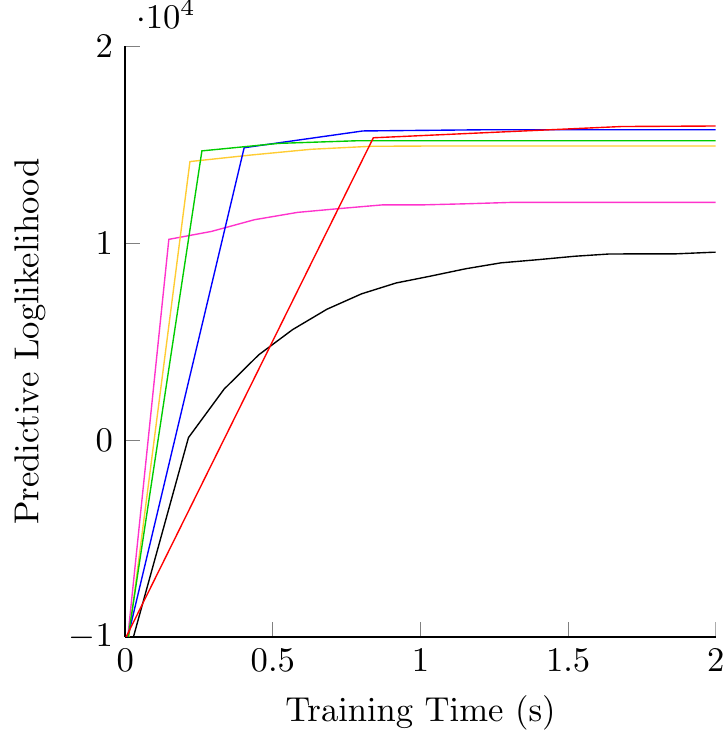}
\caption{Predicitve loglikelihood versus training time on synthetically generated data using Gibbs-SSVI with burn-in lengths of 0,1,3,5,10 and 25. Converged predictive loglikelihood is monotonically increasing in burn-in length, so no legend is included.}
\label{varyburnin}
\end{figure}

\begin{figure*}[t]
\centering
\begin{minipage}{0.55\textwidth}
   \subfigure[Gibbs initialization]{\includegraphics[trim = 98mm 185mm 47mm 20mm, clip,width=0.5\linewidth]{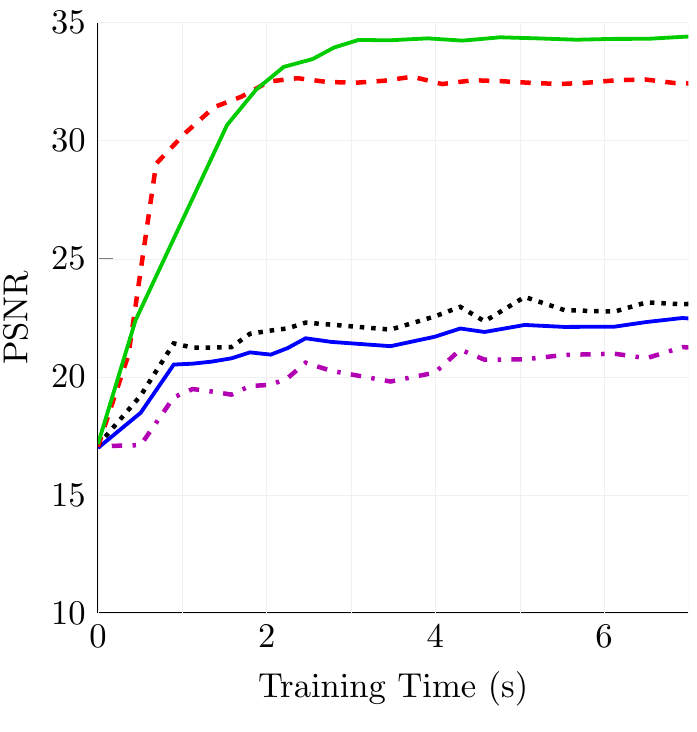}}
   \subfigure[Random initialization ]{\includegraphics[trim = 25mm 185mm 120mm 20mm, clip,width=0.5\linewidth]{boat_graphs.pdf}}
\end{minipage}
\hspace{10mm}
\begin{minipage}{0.34\textwidth}
   \centering
    \includegraphics[width=0.49\linewidth]{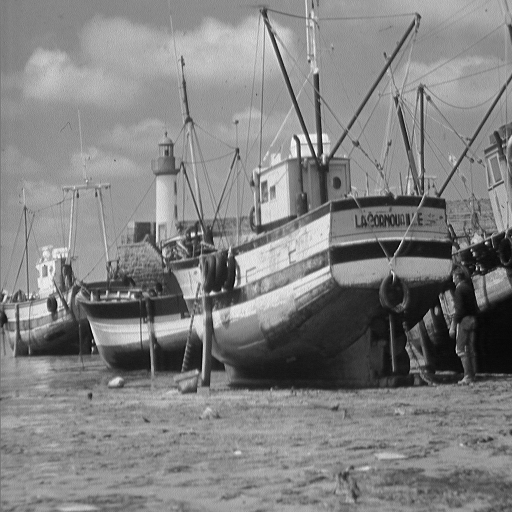}
    \includegraphics[width=0.49\linewidth]{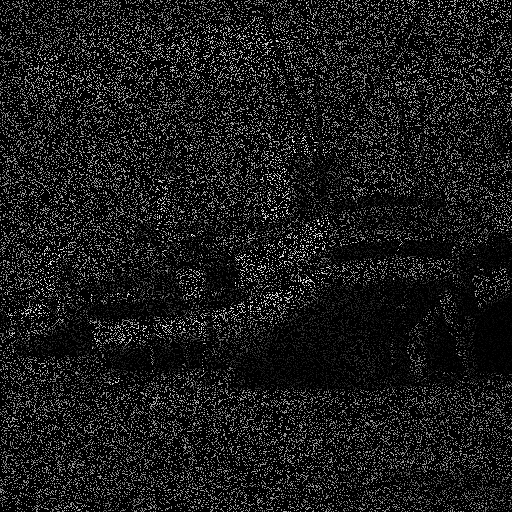} \\
   \vspace{0.8mm}
    \includegraphics[width=0.49\linewidth]{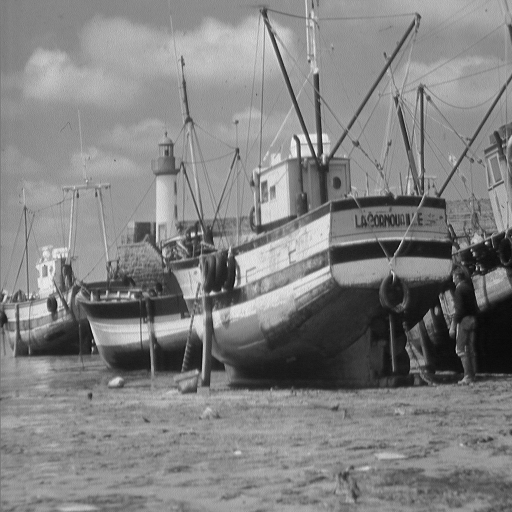} 
    \includegraphics[width=0.49\linewidth]{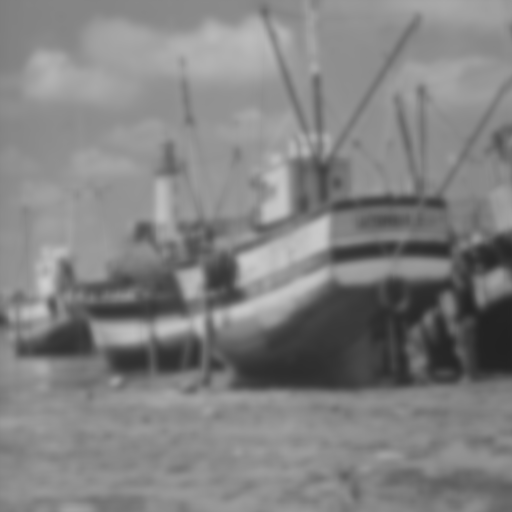} 
\end{minipage}
\caption{Results from interpolation of the $512 \times 512$ pixel `Boat' image. PSNR vs training time shown using (a) Gibbs initialization and (b) random initialization. The pictures shown are the original image (top left), the image to be reconstructed with 80\% of pixels unobserved (top right), the Gibbs-SSVI reconstruction (bottom left) and the MF-SSVI reconstruction (bottom right).  }
\label{boat}
\end{figure*}

\subsection{Synthetic Data}

A key question that is important to consider when using variational approximations of a particular form is, `how close is the approximation to the true posterior?'. We attempt to answer such a question with our first experiment. Data was generated from our prior with parameters $\gamma_w =1$, $\gamma_\mathrm{obs}=100$, $K=80$, $N=1e5$ and $D=40$, with 7.5\% selected uniformly at random held out for testing on 100 independent experiments. We then applied Gibbs-SSVI and MF-SSVI, as well as an uncollapsed Gibbs sampler to the generated data using $K=150$ potential features and random initialization. The predictive mean squared errors (MSE) of the 3 methods were $0.022 \pm 0.002$, $0.027 \pm 0.004$ and $0.020 \pm 0.002$ and the average per iteration training times were 1.5, 0.6 and 7.6 seconds respectively. Figure \ref{gibbsvsSVI} illustrates our findings. The Gibbs sampler achieves a high predictive likelihood, but the average training time per iteration was very high versus the SVI methods. 
Warm starting the Gibbs sampler at each iteration helped Gibbs-SSVI to converge in few iterations, whereas stochastically choosing subsets of data points in SSVI methods requires re-initializing local variables at each epoch. Notice that the predictive MSE of Gibbs-SSVI is close to the Gibbs sampler, suggesting that the correct mean is being learnt. The lower likelihood the SSVI methods are able to achieve is therefore due to a poor calibration in posterior variance, a known issue with variational methods \citep{consonni2007}. 

Another question of interest to us was, `what is the empirical trade-off between training time and unbiasedness in the Gibbs-SSVI scheme?'. More specifically, if we allow the Gibbs sampler over the local variables to converge, the subsequent ELBO gradient estimates would be unbiased, whilst using samples from a Gibbs chain which has not converged would lead to biased gradient estimates. Convergence of the Gibbs chain, however, may take a long time. Therefore we experimented with a range of burn-in lengths of the Gibbs chain on synthetically generated data. Various burn-in and sample length combinations were discussed by \citet{mimno2012}. We tried burn-in lengths of 0,1,3,5,10 and 25 whilst fixing the number of samples used after burn-in to 3. The results can be seen in Figure \ref{varyburnin}. When the burn-in length is below 3 we notice severe loss in predictive power of the Gibbs-SSVI method. We notice diminishing gains in predictive power as we increase the length of burn-in. This 
experiment suggests that some bias introduced by using samples from an unconverged Gibbs chain may be worth the reduction in training time. For subsequent experiments, we fix the burn-in length to 3.

\subsection{Image Interpolation and Denoising}

\citet{zhou2009} first applied the beta process for sparse image representation with good results and much follow up research. 
The standard metric used for quantifying the quality of a reconstructed image is the peak signal-to-noise ratio (PSNR), defined as $20 \log_{10} ( \mathrm{max_{image}} /\mathrm{rmse})$, where $\mathrm{max_{image}}$ is the maximum possible pixel value and $\mathrm{rmse}$ is the root mean squared error of the reconstruction. 

We consider overlapping $8 \times 8$ pixel patches as individual $64$ dimensional data points. The fact that the patches are overlapping technically breaks the exchangeability assumption of the prior distribution, however the extra model averaging is beneficial to prediction. Five grayscale images originally from \citet{portilla2003} were used for our study: 
Boat, Barbara, Lena, House and Peppers. The first 3 are $512 \times 512$ in size whilst the last 2 are $256 \times 256$. The datasets are therefore of size $N = (512-7)^2=255,025$ and $N = (256-7)^2=62,001$ for $512 \times 512$ and $256 \times 256$ images respectively.
We use a batchsize of $N_\mathrm{subset} = 250$ and $K=250$ features for our experiments.

For our first experiment, we consider the task of image interpolation, where the task is to reconstruct an image where only 20\% of the pixels, chosen uniformly at random, are observed. \citet{li2012} consider a mean field based variational approximation for such a task, however, the learning rate schedule they used was $\rho_t = (t+1000)^{-0.5}$. This implies $\rho_t<0.032$ for all $t \geq 1$, and that their algorithm relied heavily on the initialization of global parameters. They ran an MCMC algorithm over a subset of the data to initialize these global parameters, and we argue that this was integral to the performance of their algorithm. We decided to test how sensitive the variational algorithms were to different initialization methods and an example can be seen in the performance graphs of Figure \ref{boat}. We found that initializing using MCMC improved the PSNR of MF-SVI, MF-SSVI and Titsias-SSVI by 4.8 on average versus random initialization. However, the analogous improvement for Gibbs-SSVI and 
Mimno-SVI was 1.0. This suggests that the methods which preserve intra-local variable structure are less sensitive to initialization. 

\begin{figure}
\centering
\begin{minipage}{0.36\textwidth}
   \centering
    \includegraphics[width=0.425\linewidth]{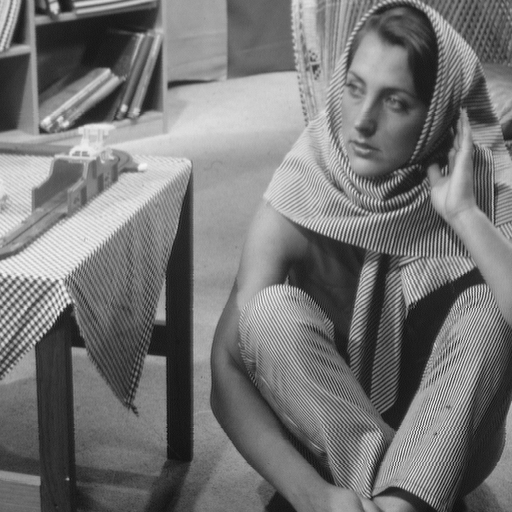}
    \includegraphics[width=0.425\linewidth]{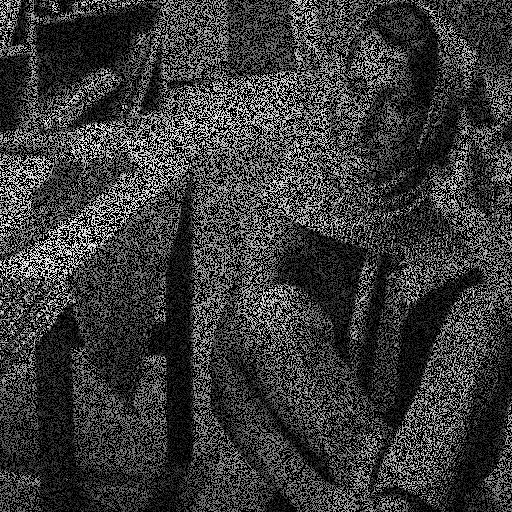} \\
   \vspace{0.8mm}
    \includegraphics[width=0.425\linewidth]{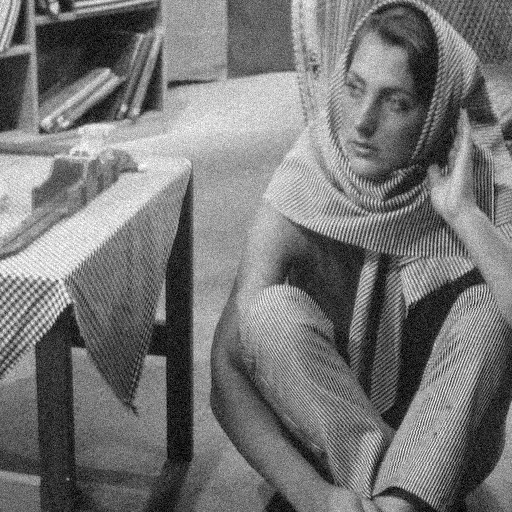} 
    \includegraphics[width=0.425\linewidth]{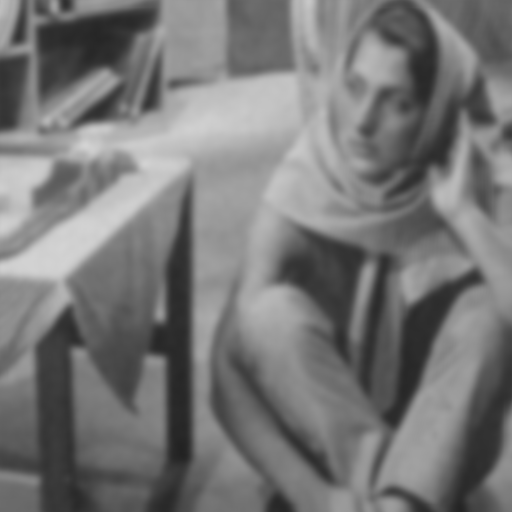} 
\end{minipage}
\caption{Results from interpolation and denoising of the $512 \times 512$ pixel `Barbara' image. The pictures shown are the original image (top left), the image to be reconstructed with 50 \% of pixels unobserved and remaining pixels corrupted with Gaussian noise (top right), the Gibbs-SSVI reconstruction (bottom left) and the MF-SSVI reconstruction (bottom right).  }
\label{barbara}
\end{figure}

Secondly, we considered the joint task of image interpolation and denoising. Here, we observe 50 \% of the pixels chosen uniformly at random, except they are now corrupted with Gaussian noise with standard deviation 15 (the original pixels take integer values in $[0,255]$). Results for both image interpolation and denoising tasks are summarized in Table \ref{image-results}. Gibbs-SSVI and Mimno-SVI consistently outperform the other methods and an explanation, outlined in Section \ref{sec:thought}, as to why this is the case can be deduced by studying the images in Figures \ref{boat} and \ref{barbara}. 

For the $512 \times 512$ images, the average training time per epoch was $0.12, 0.12, 0.13, 0.46$ and 
$0.48$ secs for the MF-SVI, MF-SSVI, Titsias-SSVI, Mimno-SVI and Gibbs-SSVI methods respectively, on a 2.4GHz dual core machine.
Among the multiple experiments, the MF-SSVI reconstructions were similar in appearance to the MF-SVI and Titsias-SSVI methods, whilst the Gibbs-SSVI reconstructions were similar to the Mimno-SVI ones. We believe the blurred appearance of the former 3 methods' reconstructions is a result of the independence between $z_{nk}$ and $z_{nk'}$ for $k \neq k'$ in their variational forms. In contrast, the Gibbs-SSVI and Mimno-SVI methods maintain dependence between $z_{nk}$ and $z_{nk'}$, and are therefore able to select a subset of features which collectively best explain the data. The latter methods are consequently much more capable of capturing structure and detail in the images we tested on, and there is a mild cost to pay in extra training time. 


\begin{figure}[!t]
  \centering
    \includegraphics[trim = 25mm 185mm 117mm 20mm, clip,width=0.8\linewidth]{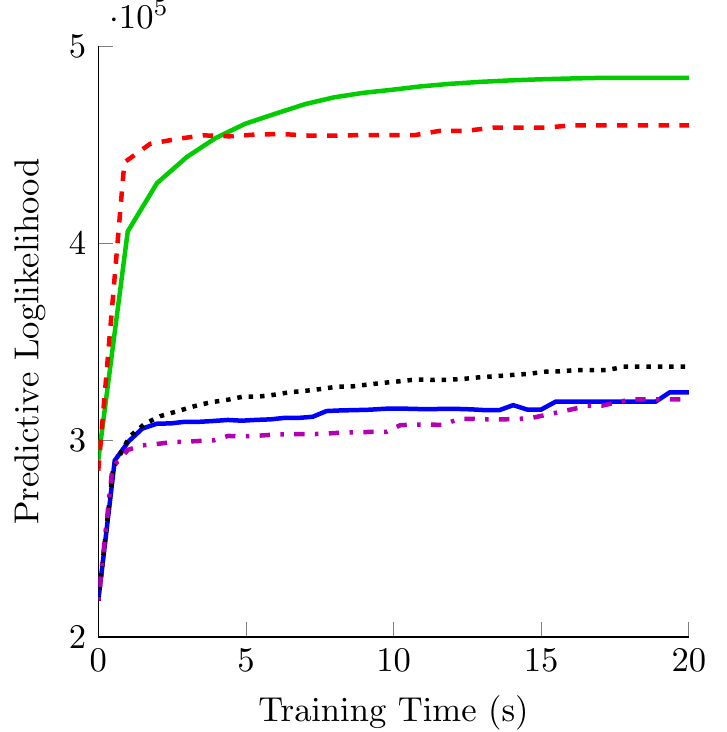}
\caption{Predicitve loglikelihood versus training time on cell line data comparing five SVI algorithms.}
\label{ccle}
\end{figure}

\begin{table*}[t]
\caption{PSNR performance of image interpolation (left entries) and denoising (right entries) tasks using Gibbs initialization of global parameters on a randomly chosen subset of data.}
\label{image-results}
\vskip 0.15in
\begin{center}
\begin{small}
\begin{sc}
\begin{tabular}{ | l |C{1cm} |C{1cm} |C{1cm} |C{1cm} |C{1cm} |C{1cm} |C{1cm} |C{1cm} |C{1cm} |C{1cm} |}
\hline
\abovespace\belowspace
 & \multicolumn{2}{c|}{Boat} & \multicolumn{2}{c|}{Barbara} & \multicolumn{2}{c|}{Lena} & \multicolumn{2}{c|}{House} & \multicolumn{2}{c|}{Peppers}\\
\hline
\abovespace
MF-SVI         & 21.1           & 19.5         & 21.8        &  20.6      & 24.1    & 23.6           & 25.3         & 24.2     & 25.9 & 24.4\\
MF-SSVI        & 22.3          & 20.8         & 22.2        &  21.4      & 24.7    &  24.4          & 26.7         & 25.4     & 25.8 & 24.1\\
Titsias-SSVI   & 23.2          & 21.5         & 22.1        &   21.7     & 26.3    & 25.8           & 26.7         & 25.3     & 27.9 & 26.8\\
Mimno-SVI    & 32.4          & 29.7         & 36.2 & 35.1  & 39.4   & 36.9          & \bf{42.8}  & \bf{40.1}     & 43.7 & 40.4\\
\belowspace
Gibbs-SSVI    & \bf{34.3}   & \bf{31.5}  & \bf{38.2}        & \bf{37.0}     & \bf{43.3} & \bf{41.7}   & 40.5       & 37.8     & \bf{47.4} & \bf{42.3}\\
\hline
\end{tabular}
\end{sc}
\end{small}
\end{center}
\vskip -0.1in
\end{table*}

\subsection{A Thought Experiment} \label{sec:thought}

To illustrate the problem with breaking dependencies between $z_{ik}$ and
$z_{ik'}$ for $k \neq k'$, we can consider a simple thought experiment. 
Suppose $\boldsymbol{y}_i = \boldsymbol{f} + \boldsymbol{\epsilon}_i$, 
$\boldsymbol{y}_i \in \mathbb{R}^D$, where 
$\boldsymbol{f} \sim \mathcal{N}(0,\boldsymbol{I})$ and 
$\boldsymbol{\epsilon}_i \sim \mathcal{N}(0,0.05 \boldsymbol{I})$ independently for $i=1,...,N$.
Now consider applying MF-SSVI and Gibbs-SVI algorithms to this dataset, whilst fixing $w_{ik} = 1$ 
for each $i,k$, and using $K=2$ features.
Let's assume that at the current iteration $\pi_1^{(t)} \approx \pi_2^{(t)}$ and 
$\boldsymbol{\phi}_1^{(t)} \approx \boldsymbol{\phi}_2^{(t)} \approx \boldsymbol{f}$. 
The local ELBO for the MF-SSVI method will have local optima for $\theta_{i1} = 1-\theta_{i2}$,
since exactly 1 feature is needed to explain the data, so MF-SSVI would find a local optimum of the
form $q(z_{i1}) = \mathrm{Bernoulli}(s)$, $q(z_{i2}) = \mathrm{Bernoulli}(1-s)$ for some
$s \in (0,1)$. (We were able to verify this form of local optimum empircally). The Gibbs-SSVI will generate
samples of the form $\boldsymbol{z}_i = (0,1)$ and $\boldsymbol{z}_i = (1,0)$.  

We would like to  
predict $\boldsymbol{y}_i$ with each model. Since $q(z_{i1})$ and $q(z_{i2})$ are independent
under MF-SSVI, predictions will be $\hat{\boldsymbol{y}}_i = 0$ with probability $s(1-s)$, 
$\hat{\boldsymbol{y}}_i \approx \boldsymbol{y}_i$ with probability $s^2+(1-s)^2$ and finally
$\hat{\boldsymbol{y}}_i \approx 2\boldsymbol{y}_i$ with probability $s(1-s)$. Conversely, 
the Gibbs samper in Gibbs-SSVI will place little to no probability on both, or neither
of the 2 features being used for prediction. In summary, the Gibbs based local variable estimates 
can handle the strong correlation between the 2 features, whilst the MF based method cannot 
and suffers dramatically because of it. 

One possible solution to this problem would be to encourage all features to have limited correlation
apriori, discouraging situations where multiple features are learned to be similar to each other. 
Encouraging dissimilarity in such a way is challenging and would complicate the
otherwise clean updates that are possible in SVI methods. 

\vspace{-1mm}
\subsection{Genomic data}

Vast amounts of genomic data are currently being collected as technology advances. It will be crucial to develop machine learning models and more importantly, inference algorithms, which can cope with large data sets, whilst still retaining flexible modelling ability. We consider 2 datasets for which sparse latent feature modelling is appropriate. We use $K=500$ features for both experiments. 

\vspace{-3mm}
\paragraph{Cancer cell line data.}  The Cancer Cell Line Encyclopedia is a collection of around 450
cancer samples including gene expression, copy number variation, and
drug response information. We focus on modeling the gene expression
data, which has measurements for around 15,000 genes. In this setting
we are more interested in finding overlapping clusters (sparse
features) of \emph{genes} rather than samples, so we effectively have
$N=15000, D=450$. The latent factors found can then be
interpreted as biological pathways, or sets of genes regulated by the
same transcription factor. Understanding the structure in this data is
valuable as a first step towards associating the cellular
characteristics of the cancers to their drug response profiles.
We randomly hold out 10 \% of the data for testing. 
Results for this experiment are summarized in Figure \ref{ccle}.

\vspace{-3mm}
\paragraph{CyTOF data. }  CyTOF is a novel extremely high through-put technology capable of
measuring up to 40 protein abundance levels in thousands of individual
cells per second. The cells are controlled using flow cytometry and
specific proteins are tagged using  heavy metals
which can be measured using time-of-flight mass spectrometry. Existing
analyses have attempted to group the observed cells into
non-overlapping subpopulations, but we here show that the data can be
effectively modeled as compromising of a spectrum of cell types
expressing different latent factors to differing extents. The sample
we analyse consists of human immune cells, so representing the
heterogeneity is relevant for understanding disease response. Our dataset has $N=532,000, 
D=40$ and a random 5\% is used as test data. 
The results for the experiments on this data follow a very similar pattern to that of the cell
line gene expression data. The converged predictive log-likelihoods after training for 10 minutes
are $-1.1e6$, $-9.6e5$, $-9.4e5$, $-3.8e5$ and $-3.2e5$ 
for the MF-SVI, MF-SSVI, Titsias-SSVI, Mimno-SVI and Gibbs-SSVI methods
respectively.

\vspace{-1mm}
\section{Conclusions}

In this work, we compare various stochastic variational inference algorithms for beta process factor analysis.
Whist many methods in the literature have been proposed, we have chosen to exploit the conditional conjugacy
and the exponential family nature of our model to create simple natural parameter updates.

\citet{ssvi} found that preserving structure between local and global variables significantly boosted 
performance for the LDA, but based on our experiments, we conclude that preserving intra-local variable dependence is crucial to 
prediction in the beta-Bernoulli process. This is evident from the fact that both Gibbs-SSVI and Mimno-SVI
consistently and significantly outperform MF-SVI, MF-SSVI and Titsias-SSVI on a variety of image interpolation and denoising 
tasks and on modelling genomic data. The Titsias-SSVI method models dependence between $z_{ik}$ and $w_{ik}$, but does not appear to significantly outperform MF-SSVI, suggesting that this dependence is not crucial in prediction. 
Mimno-SVI does not maintain dependence between local and global variables whilst MF-SSVI does, and yet Mimno-SVI leads to better predictions. 
We discuss why this is the case in a simple thought experiment, showing the benefit of maintaining dependence between local variables where $k\neq k'$. 
The multi-cluster, sparse nature of the beta-Bernoulli process makes mean field type local variable approximations highly sensitive to correlated features. 
Gibbs-SSVI does also modestly outperform Mimno-SVI through maintaining dependencies between global and local variables. 

In summary, care is needed to ensure that the dependencies encoded by a particular variational
approximation are appropriate for the model being considered.  
\bibliography{ref}
\bibliographystyle{icml2015}

\newpage
\onecolumn

\section*{Appendix}

Here, we provide details about the local variable approximations introduced in the main text of the paper. 

\subsection*{Mimno-SVI}

The form of the local approximation in the Mimno-SVI method is
\begin{align*}
\log q_\mathrm{Mimno}(\boldsymbol{\psi}_i) &=  \mathbb{E}_{q(\boldsymbol{\beta})}[\log p(\boldsymbol{\psi}_i|\boldsymbol{y}_{1:N},\boldsymbol{\beta})]  \notag \\
& = \mathbb{E}_{q(\boldsymbol{\beta})} \bigg[
-\frac{\gamma_\mathrm{obs}}{2} \big\| \boldsymbol{y}_i - (\boldsymbol{z}_i  \circ \boldsymbol{w}_i ) \boldsymbol{\Phi}   \big\|^2  +  \sum_k z_{ik} \log \bigg( \frac{\pi_k}{1-\pi_k} \bigg) - \frac{\gamma_w}{2} \boldsymbol{w}_i \boldsymbol{w}_i^\top
\bigg] + \mathrm{const} \\
&= -\frac{c}{2d}  \sum_k z_{ik} w_{ik} \bigg[ w_{ik}  
\bigg(\frac{\boldsymbol{\mu}_k \boldsymbol{\mu}_k^\top}{{\tau_k}^2} + \frac{1}{\tau_k}\bigg) 
+ \bigg( \sum_{j \neq k} z_{ij} w_{ij} \frac{\boldsymbol{\mu}_k \boldsymbol{\mu}_j^\top}{\tau_k \tau_j}\bigg)  
- 2 \frac{\boldsymbol{\mu}_k \boldsymbol{y}_i^\top}{\tau_k}
\bigg] \\
& \hspace{8mm} + \sum_k z_{ik} (\psi(a_k)-\psi(b_k))  -\frac{e}{2f} 
\boldsymbol{w}_i \boldsymbol{w}_i^\top  + \mathrm{const}
\end{align*}
It is clear that $\log q_\mathrm{Mimno}$ is quadratic in each $w_{ik}$ and linear in each $z_{ik}$, therefore a Gibbs based sampler can easily be consructed to sample from $q_\mathrm{Mimno}$, where $w_{ik}$ is Gaussian given all other local variables, and $z_{ik}$ is Bernoulli given all other local variables.

\subsection*{MF-SSVI}

The local ELBO in the MF-SSVI framework is very similar to that of the MF-SVI, the difference being that samples of the global variables are used in MF-SSVI. The local ELBO has the following form

\begin{align*}
\mathcal{L}_\mathrm{local}^\mathrm{MF-SSVI} &= \frac{\gamma_\mathrm{obs}}{2} \sum_{i,k}\theta_{ik} \boldsymbol{\phi}_k
 \bigg[ 2 \frac{\nu_{ik}}{\kappa_{ik}}  \boldsymbol{y}_i^\top 
 - \bigg(\frac{{\nu_{ik}}^2}{{\kappa_{ik}}^2} + \frac{1}{\kappa_{ik}} \bigg)
\boldsymbol{\phi}_k^\top 
 - \sum_{j \neq k} \theta_{ij} \frac{\nu_{ij}}{\kappa_{ij}} \frac{\nu_{ik}}{\kappa_{ik}}
\boldsymbol{\phi}_j^\top \bigg] \\
 &\hspace{10mm}- \frac{\gamma_w}{2} \sum_{i,k}
\bigg(\frac{{\nu_{ik}}^2}{{\kappa_{ik}}^2} + \frac{1}{\kappa_{ik}} \bigg)  + \sum_{i,k} \theta_{ik} \bigg( \frac{\pi_k}{1-\pi_k} \bigg) \\
&\hspace{10mm}- \frac{1}{2} \sum_{i,k} \log (\kappa_{ik}) \notag  - \sum_{i,k} \big[ \theta_{ik} \log \theta_{ik} + (1-\theta_{ik}) \log (1-\theta_{ik})  \big].
\end{align*}

This is optimized as a function of $\{\theta_{ik}, \nu_{ik}, \kappa_{ik}$ using gradient descent. Once a local optimum is found, $\mathbb{E}_{q_\mathrm{MF}(\boldsymbol{\psi}_{1:N}|\boldsymbol{\beta}^{(t)})}[\boldsymbol{\eta}_i]$ can be computed analytically as a function of the optimized parameters and global variable samples.  

\subsection*{Titsias-SSVI}

Recall that the Titsias-SSVI method maintains dependence between $z_{ik}$ and $w_{ik}$ for each $k$. The local ELBO for Titsias-SSVI is

\begin{align*}
\mathcal{L}_\mathrm{local}^\mathrm{Titsias-SSVI} &= \frac{\gamma_\mathrm{obs}}{2} \sum_{i,k}\theta_{ik} \boldsymbol{\phi}_k
 \bigg[ 2 \frac{\nu_{ik}}{\kappa_{ik}}  \boldsymbol{y}_i^\top 
 - \bigg(\frac{{\nu_{ik}}^2}{{\kappa_{ik}}^2} + \frac{1}{\kappa_{ik}} \bigg)
\boldsymbol{\phi}_k^\top 
 - \sum_{j \neq k} \theta_{ij} \frac{\nu_{ij}}{\kappa_{ij}} \frac{\nu_{ik}}{\kappa_{ik}}
\boldsymbol{\phi}_j^\top \bigg] \\
 &\hspace{10mm}- \frac{\gamma_w}{2} \sum_{i,k}
\bigg(\frac{{\nu_{ik}}^2}{{\kappa_{ik}}^2} + \frac{1}{\kappa_{ik}} \bigg)  + \sum_{i,k} \theta_{ik} \bigg( \frac{\pi_k}{1-\pi_k} \bigg) \\
&\hspace{10mm}- \frac{1}{2} \sum_{i,k}\Big[ \theta_{ik}
\big( \log (\kappa_{ik} ) - 1 \big)
+(1-\theta_{ik})
\big( \log (\gamma_w ) - 1 \big)
\Big]\\
&\hspace{10mm} - \sum_{i,k} \big[ \theta_{ik} \log \theta_{ik} + (1-\theta_{ik}) \log (1-\theta_{ik})  \big].
\end{align*}

Again, this function is maxized as a function of $\{\theta_{ik}, \nu_{ik}, \kappa_{ik}$ using gradient descent, and the optimized parameters along with the global variable samples are used to compute $\mathbb{E}_{q_\mathrm{Titsias}(\boldsymbol{\psi}_{1:N}|\boldsymbol{\beta}^{(t)})}[\boldsymbol{\eta}_i]$ analytically.

\subsection*{Gibbs-SSVI}

The Gibbs-SVI method uses the true posterior conditional distribution for local variables
\begin{align*}
\log q_\mathrm{Gibbs}(\boldsymbol{\psi}_i) &=  \log p(\boldsymbol{\psi}_i|\boldsymbol{y}_{1:N},\boldsymbol{\beta})  \notag \\
& = -\frac{\gamma_\mathrm{obs}}{2} \big\| \boldsymbol{y}_i - (\boldsymbol{z}_i  \circ \boldsymbol{w}_i ) \boldsymbol{\Phi}   \big\|^2  +  \sum_k z_{ik} \log \bigg( \frac{\pi_k}{1-\pi_k} \bigg) - \frac{\gamma_w}{2} \boldsymbol{w}_i \boldsymbol{w}_i^\top + \mathrm{const} \\
&= -\frac{\gamma_\mathrm{obs}}{2}  \sum_k z_{ik} w_{ik} \boldsymbol{\phi}_k \bigg[ w_{ik}  
 \boldsymbol{\phi}_k^\top 
+ \bigg( \sum_{j \neq k} z_{ij} w_{ij}  \boldsymbol{\phi}_j^\top \bigg)  
- 2 \boldsymbol{y}_i^\top
\bigg] \\
& \hspace{8mm} +  \sum_k z_{ik} \log \bigg( \frac{\pi_k}{1-\pi_k} \bigg) - \frac{\gamma_w}{2} \boldsymbol{w}_i \boldsymbol{w}_i^\top + \mathrm{const} 
\end{align*}

Just as was the case with Mimno-SVI, we notice that $\log q_\mathrm{Gibbs}$ is quadratic in each $w_{ik}$ and linear in each $z_{ik}$, therefore a Gibbs sampler can be designed to sample from $q_\mathrm{Gibbs}$.

\end{document}